\documentclass[sigconf]{acmart}

\usepackage{algorithm}
\usepackage{algorithmic}
\usepackage{multirow}
\usepackage{makecell}
\usepackage{tabularx}
\usepackage{bbding}
\usepackage{tabularray}
\usepackage{balance}

\AtBeginDocument{%
  }

\copyrightyear{2025} 
\acmYear{2025} 
\setcopyright{acmlicensed}\acmConference[MM '25]{Proceedings of the 33rd ACM International Conference on Multimedia}{October 27--31, 2025}{Dublin, Ireland}
\acmBooktitle{Proceedings of the 33rd ACM International Conference on Multimedia (MM '25), October 27--31, 2025, Dublin, Ireland}
\acmDOI{10.1145/3746027.3758211}
\acmISBN{979-8-4007-2035-2/2025/10}

\begin{document}

\title{SmokeBench: A Real-World Dataset for Surveillance Image Desmoking in Early-Stage Fire Scenes}

\author{Wenzhuo Jin}
\authornote{Both authors contributed equally to this research.}
\orcid{0009-0005-3413-8723}
\affiliation{%
  \institution{Beijing Jiaotong University}
  \city{Beijing}
  \country{China}
}
\email{23722011@bjtu.edu.cn}

\author{Qianfeng Yang}
\authornotemark[1]
\affiliation{%
  \institution{Dalian Polytechnic University}
  \city{Dalian}
  \country{China}
  }
\email{csqianfengyang@163.com}

\author{Xianhao Wu}
\authornotemark[1]
\affiliation{%
  \institution{Nanjing University of Aeronautics and Astronautics}
  \city{Nanjing}
  \country{China}
  }
\email{wuxianhao@nuaa.edu.cn}

\author{Hongming Chen}
\affiliation{%
  \institution{Dalian Martime University}
  \city{Dalian}
  \country{China}
  }
\email{chenhongming@dlmu.edu.cn}

\author{Pengpeng Li}
\affiliation{%
  \institution{Nanjing University of Science and Technology}
  \city{Nanjing}
  \country{China}
  }
\email{pengpengli@njust.edu.cn}

\author{Xiang Chen}
\affiliation{%
  \institution{Nanjing University of Science and Technology}
  \city{Nanjing}
  \country{China}
  }
\email{chenxiang@njust.edu.cn}






\renewcommand{\shortauthors}{Wenzhuo Jin et al.}

\begin{abstract}
Early-stage fire scenes (0-15 minutes after ignition) represent a crucial temporal window for emergency interventions.
During this stage, the smoke produced by combustion significantly reduces the visibility of surveillance systems, severely impairing situational awareness and hindering effective emergency response and rescue operations.
Consequently, there is an urgent need to remove smoke from images to obtain clear scene information.  
However, the development of smoke removal algorithms remains limited due to the lack of large-scale, real-world datasets comprising paired smoke-free and smoke-degraded images.
To address these limitations, we present a real-world surveillance image desmoking benchmark dataset named  {\bfseries SmokeBench}, which contains image pairs captured under diverse scenes setup and smoke concentration. 
The curated dataset provides precisely aligned degraded and clean images, enabling supervised learning and rigorous evaluation.
We conduct comprehensive experiments by benchmarking a variety of desmoking methods on our dataset.
Our dataset provides a valuable foundation for advancing robust and practical image desmoking in real-world fire scenes.
This dataset has been released to the public and can be downloaded from~\href{https://github.com/ncfjd/SmokeBench} {\textit{https://github.com/ncfjd/SmokeBench}}.
\end{abstract}




\maketitle

\section{Introduction}
\begin{figure}[t]
  \centering
  \includegraphics[width=\linewidth]{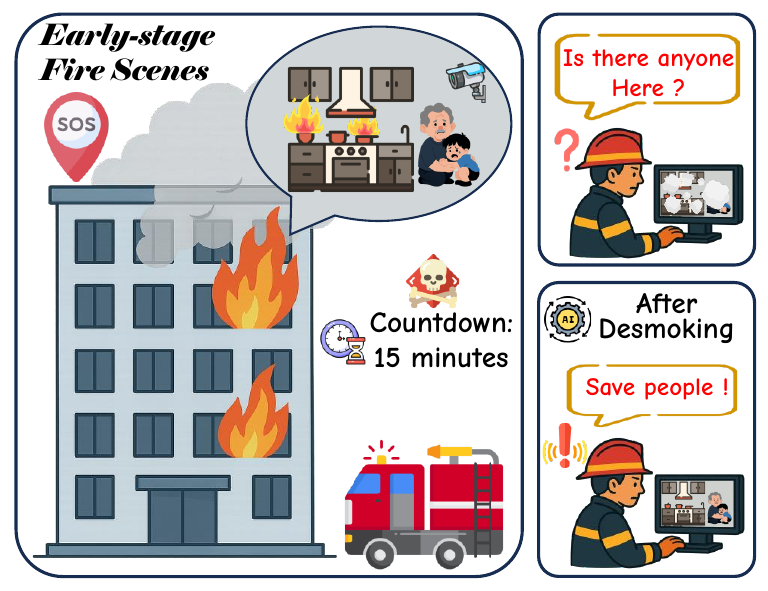}
  \caption{Illustration of emergency response in early-stage fire scenes.
  Dense combustion smoke degrades surveillance visibility, impairing situational awareness and hindering rescue operations. 
  Paired real-world smoke-free/smoke-degraded image datasets are essential for developing effective desmoking algorithms to support fire emergency response.}
  \label{Fig1}
\end{figure}

Early-stage fire scenes~\cite{hine2004fire} specifically refer to the initial 15-minute period following fire ignition, which represents a critical phase in fire development. 
During this stage, flames are typically confined to the vicinity of the ignition source with limited spread range and relatively low heat release rate. 
As the fire has not yet fully developed, the smoke layer remains incompletely settled, maintaining some visibility at the scene.
These conditions provide favorable circumstances for occupant evacuation and initial fire suppression efforts. 
This phase is therefore regarded as the "golden window period" for fire incident management.
Failure to effectively control the fire during this early stage will lead to its progression into a rapid growth phase after 15 minutes. 
During this subsequent stage, the heat release rate increases dramatically, with fire intensity growing exponentially and potentially triggering hazardous phenomena such as flashover. 
Thus, fully leveraging the early-stage window is vital for minimizing escalation and maximizing rescue success.

\begin{figure*}[t]
    \centering
    \begin{tabular}{c}
        \includegraphics[width=\linewidth]{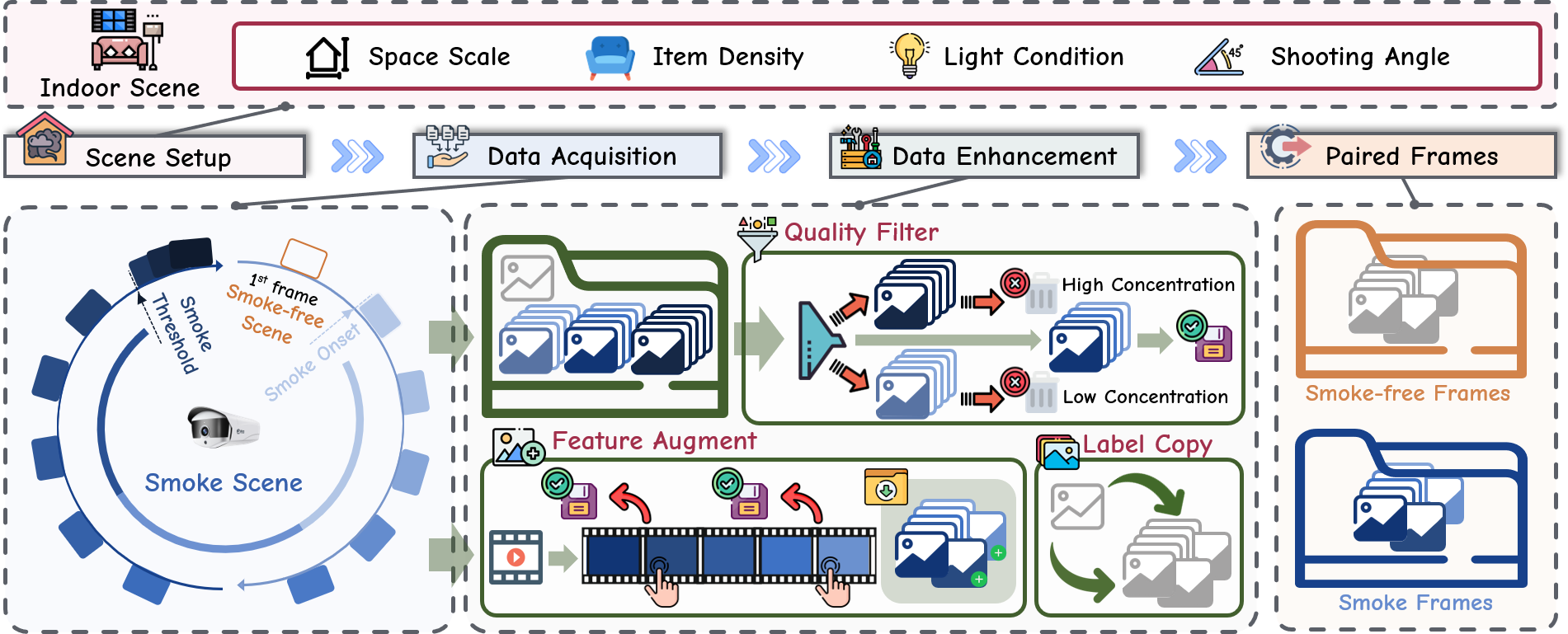}
    \end{tabular}
    \caption{Illustration of the early-stage fire scenes smoke acquisition system (SAS). SAS consists of three modules: Scene Setup, Data Acquisition, and Data Enhancement.
    It constructs diversified fire scenes based on space scale, item density, light condition and shooting angle, employing video recording to capture smoke diffusion processes. 
    Through quality filter, feature augment, and label copy, it ultimately generates precisely aligned clean and smoke-degraded image pairs for model training.}
    \label{fig:data_pipeine}
\end{figure*}
%
During this critical period, indoor surveillance systems often remain fully operational.
Effective utilization of surveillance systems could provide vital information support for fire rescue operations, as illustrated in Figure~\ref{Fig1}.
Nevertheless, a critical operational challenge emerges in these circumstances.
While surveillance equipment typically remains undamaged by flames during this initial phase, the smoke produced by combusting materials rapidly disseminates throughout the space, significantly degrading surveillance visibility~\cite{kodur2020fire,mclauchlan2020fire}. 
Such visual interference directly impacts firefighters' ability to quickly and accurately locate the fire origin, while substantially reducing the efficiency of identifying the specific floors where trapped occupants may be located.
Therefore, developing techniques for surveillance image desmoking during early-stage fire scenarios holds considerable practical importance.
Recently, the introduction of new techniques from machine learning and deep learning provides a broader perspective for image processing~\cite{chen2023learning, chen2024bidirectional, song2023image,song2023vision,jiang2023dawn,zhong2022rainy,yang2023good,zhong2023refined,chen2022unpaired,chen2025towards,li2024foundir}.
However, compared to the extensively studied image dehazing task~\cite{song2023learning, fattal2008single, ren2016single, engin2018cycle,liu2021synthetic,li2021dehazeflow,zhang2020nighttime}, image desmoking remains significantly overlooked. 
This neglect primarily stems from the research's frequent conflation of smoke removal with haze removal. 
In reality, smoke~\cite{chu2022physics} and haze~\cite{ancuti2020nh} differ fundamentally in their physical origins and visual manifestations~\cite{huang2021single, yamaguchi2017video}.
Specifically, haze consists of water droplets suspended in air, exhibiting relatively uniform and stable distribution without drastic local concentration variations~\cite{ cai2016dehazenet,tang2014investigating}. 
In contrast, smoke originates from carbon particles and aerosols generated by combustion, which, due to thermal convection effects, demonstrate dynamic motion characteristics and non-uniform spatial distribution patterns, often causing severe local occlusion in images~\cite{pei2019single}.
Consequently, haze removal approaches often fail to generalize to smoke-filled fire scenes, where conditions are more complex.
Moreover, the challenges of smoke removal are compounded by the limitations of image acquisition in real-world surveillance environments. 
Due to practical constraints on deployment cost and hardware capabilities, most fire monitoring systems rely on low-resolution cameras that produce images with limited detail and poor dynamic range.
These characteristics differ significantly from the high-quality RGB images frequently used in algorithm development, resulting in a performance gap when models are transferred from synthetic to real-world scenarios.
Another pressing limitation lies in the training data itself.
Existing desmoking algorithms are developed and validated using synthetic datasets~\cite{huang2021single}, generated via 3D rendering platforms such as Blender~\cite{hess2013blender}. 
While such tools facilitate controlled data generation, they typically rely on simplified parameter configurations (\textit{e.g.}, the intensity, density and position of smoke generation) and struggle to accurately simulate the complex, thermally driven behaviors of real smoke.
These discrepancies hinder the generalization capability and practical effectiveness of current smoke removal algorithms in real fire rescue scenarios.
To this end, we propose an early-stage fire scenes smoke acquisition system (SAS) specifically designed to capture real-world smoke-free/smoke-degraded surveillance image pairs.
Notably, our dataset represents the first real-world collection encompassing diverse scenes setup and smoke concentration in authentic early-stage fire scenes.
To validate the practical value of our dataset, we conduct comprehensive experiments using some methods to evaluate their performance.
We believe our work represents a significant step toward improving real-time situational awareness and response effectiveness in practical fire emergency management systems.
The contributions of our work are summarized as follows:
\begin{itemize}
\item We introduce the first real-world surveillance image desmoking benchmark dataset that includes diverse scenes setup and smoke concentration.

\item We design an early-stage fire scenes smoke acquisition system (SAS), enabling the collection of smoke-degraded and smoke-free image pairs under realistic conditions.

\item We conduct a comprehensive study and evaluation of different methods using our proposed benchmark dataset.
\end{itemize} 

\section{Related Work}
\subsection{Smoke Datasets}
%
Image desmoking is essential for firefighting navigation and surveillance, enabling the recovery of smoke-degraded images for safety-critical decisions.  Compared to dehazing, it is more challenging due to smoke’s finer particles, stronger occlusion, and dynamic, irregular dispersion, especially in confined environments.

Recently, deep learning-based desmoking methods have emerged, which rely on large-scale datasets of smoke and corresponding clean images. However, collecting such real pairs is labor-intensive due to the need for consistent illumination and static scenes.
To address this, Huang \textit{et al.}~\cite{huang2021single} propose an advanced synthetic pipeline based on the Blender, generating approximately 6,000 image pairs using physically-based smoke motion models. Their approach simulates dynamic smoke with independently controlled RGB and depth channels, incorporating variations in density, intensity, and spatial distribution. Additionally, the dataset includes 2,400 real smoke images collected from online sources and custom captures.

While synthetic data has advanced through improved rendering techniques, it still fails to capture key physical characteristics of real fire smoke.  More importantly, surveillance systems are constrained by their hardware characteristics (including low resolution, narrow dynamic range, and limited color depth) and often operate in IR night vision mode, which significantly differs from high-quality smoke images obtained from the internet or professional cameras.
These fundamental differences in imaging conditions lead to current algorithms' insufficient capability to recognize low-resolution smoke features and poor image restoration performance in night vision mode during real fire rescue scenarios, limiting the practical application value of desmoking technology in critical tasks such as firefighting command and personnel search and rescue.

\subsection{Single Image Desmoking}
Early desmoking methods are based on the atmospheric scattering model, recovering clear images by estimating transmission maps and atmospheric light. Inspired by dehazing techniques like the dark channel prior~\cite{he2010single} and color line prior~\cite{fattal2014dehazing}, they apply optical priors to constrain the degradation process. However, these methods often assume uniform smoke and single-color lighting, limiting their effectiveness in complex real-world scenes with heterogeneous smoke and diverse illumination.

With the rapid development of deep learning, end-to-end neural networks have become mainstream for single image desmoking.
CNN-based models often use multi-scale feature extraction to directly map smoky images to clear counterparts. For example, DesmokeNet~\cite{chen2021desmokenet} adopts a hierarchical structure to remove smoke of varying densities, enhancing image clarity and detail. It introduces thickness-aware pixel loss and dark channel loss to suppress residual smoke and color distortion, while self-attention and contrastive regularization further improve performance.

Moreover, generative adversarial networks (GANs) have been used in desmoking task to enhance perceptual realism via adversarial loss, proving effective in cases of dense smoke or drastic illumination changes.
DesmokeGAN~\cite{huang2021single} introduces visual attention to improve the generator's ability to capture smoke features and context, while the PatchGAN-based discriminator leverages a multi-component loss on local patches to boost desmoking quality.

Recently, Transformer-based methods have been introduced to desmoking due to their strengths in modeling long-range dependencies and global context. Models like DehazeFormer~\cite{song2023vision} leverage self-attention mechanisms to enhance non-local feature modeling across widespread smoke regions, effectively mitigating CNNs’ limitations in capturing global information.

\begin{figure*}[t]
    \centering
    \setlength{\tabcolsep}{1mm} 
    \begin{tabular}{ccc}
        \includegraphics[width=0.21\linewidth]{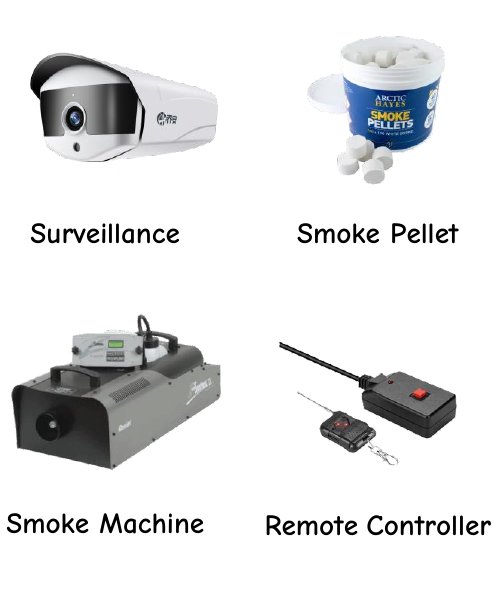} & 
        \includegraphics[width=0.36\linewidth]{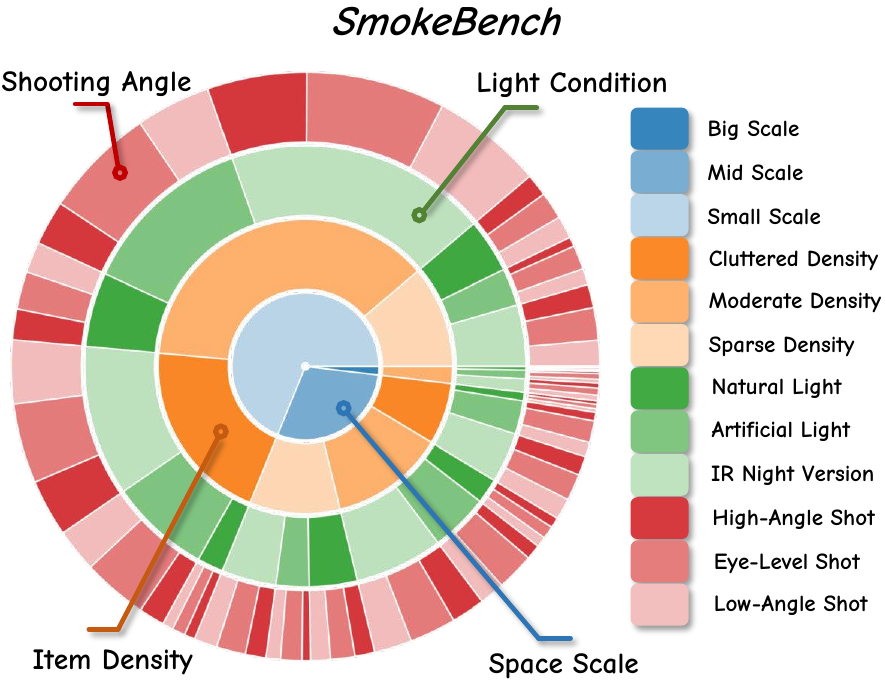} & 
        \includegraphics[width=0.41\linewidth]{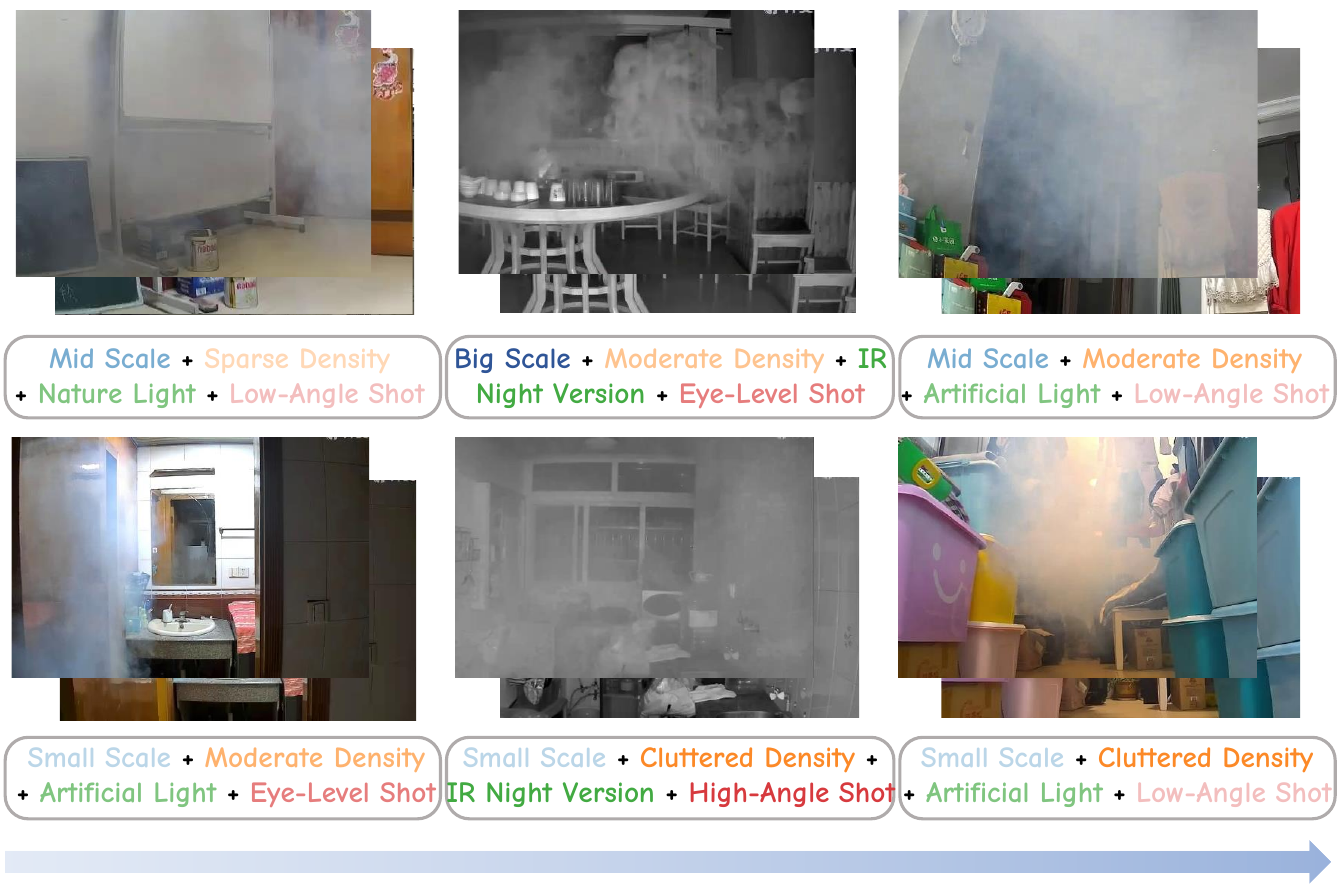} \\
        (a) Equipment  & (b) Distribution & (c) Samples
    \end{tabular}
    \caption{Illustration of the proposed SmokeBench dataset.~(a) Overview of the data collection equipment.~(b) Data distribution of the proposed dataset.~(c) Sample image pairs from our dataset. In each pair, the left image shows the LQ, while the right image displays the corresponding GT image.~We collect images with different smoke concentration under diverse
    scenes setup.}
    \label{fig:data_show}
\end{figure*}
\section{SmokeBench  Dataset}
%
To enable supervised desmoking learning, we propose a smoke imaging model and a paired data acquisition system that captures smoke and corresponding smoke-free images of the same fire scene.  Details are provided in the following sections.

\subsection{Smoke Imaging Model for Fire Scene}

In general, the atmosphere scattering model (ASM) based on radiative transfer equation (RTE) is given:
\begin{equation}
		I(x) = J(x) {\cdot} t(x) + A {\cdot} (1 - t(x)), 
\end{equation}
where $I(x)$ denotes the observed intensity at pixel $x$, $J$ denotes clear scene radiance, $t$ denotes medium transmission, and $A$ denotes global atmospheric light.
This model is widely used to dehaze in an outdoor environment with strong atmosphere scattering.


However, in indoor fire smoke scenes, visual effects differ significantly from outdoor haze due to the distinct physical properties of smoke and environmental conditions. Smoke behavior is largely influenced by particle size, confined space, and lighting. Our proposed smoke imaging model, based on the RTE with adjusted ASM parameters, treats smoke as an optically thick medium exhibiting both scattering and absorption—ranging from localized to dense, scene-wide smoke. The model excludes flaming combustion and assumes illumination is either external or absent.
%

Due to constrained indoor dimensions and short camera-to-wall distances, $A$ contributes minimally in smoke-filled scenes compared to haze, and is thus omitted. The smoke-scene image is modeled as the sum of direct scene radiance and smoke-scattered light. The smoke imaging equation for fire scenes is formulated as:
\begin{equation}
		\hat{I}(x) = \hat{J}(x) {\cdot} T(x) + S {\cdot} (1 - T(x)), 
\end{equation}
where $\hat{I}(x)$ is smoky image intensity at pixel $x$, $\hat{J}$ is clear scene radiance, $T$ is smoke transmission, and $S$ is smoke scattered light.

According to the Beer-Lambert law, transmission light through the early-stage fire smoke is modeled as:
\begin{equation}
    T(x) = e^{-{{{\beta}_e}(x)} {\cdot} d(x)},
\end{equation}
where ${\beta}_e$ denotes extinction coefficient, and $d$ denotes scene depth.
It indicates that the intensity of light reaching the camera decreases exponentially with ${\beta}_e$ and $d(x)$.
In the smoke region, the distance from the pixel $x$ and its surrounding pixels to the camera is nearly constant.
The light attenuation is described by ${\beta}_e$, which is the sum of the scattering coefficient ${{\beta}_s}$ and the absorption coefficient ${{\beta}_a}$. Scattering is the predominant cause of image degradation in smoke under visible light, with absorption contributing minimally. Thus, we disregard the absorption coefficient:
\begin{equation}
    {\beta}_e = {{\beta}_s} + {{\beta}_a} {\approx} {{\beta}_s}, 
\end{equation}

In the early stage of fire smoke, the diameter of the smoke particles increases approximately 100-300 nm~\cite{E2012the, David2012Particle}. The diameter lies within the valid range for accurate Mie scattering:
\begin{equation}
	{{{\beta}_s}(x)} = \int_0^{\infty} {N(r,x) {\cdot} Q_{sca}(r,{\lambda}) {\cdot} {\pi}{r^2}dr}, 
\end{equation}
where $N(r,X)$ denotes particle size distribution at location $x$ and with diameter $r$, $Q_{sca}(r,{\lambda})$ denotes scattering efficiency for particles of radius $r$ and wavelength $\lambda$, and ${\pi}r^2$ denotes geometric cross-sectional area.
${\beta}_e$ can be derived from the smoke concentration obtained using a smoke detector:
\begin{equation}
	{{\beta}_e}(x) = {C(x)}{\frac{{\beta}_0}{C_0}}, 
\end{equation}
where ${\beta}_0$ denotes the reference extinction coefficient, typically based on known smoke properties, $C(x)$ denotes the measured smoke concentration at location $x$, and $C_0$ the reference concentration corresponding to ${\beta}_0$.

Additionally, the texture of smoke appears to be more sharp than in outdoor scenarios due to the proximity of smoke targets to the camera and the simple backgrounds. To accurately reflect fire smoke turbulence, the scattering coefficient is further adjusted:
\begin{equation}
	{{\beta}_s}(x) = {{\beta}_0} + {\Delta}{{\beta}_{low}}(x) + {\Delta}{{\beta}_{high}}(x),
\end{equation}
where ${\beta}_0$ denotes baseline scattering coefficient, ${\Delta}{{\beta}_{low}}(x)$ is low-frequency turbulence, and ${\Delta}{{\beta}_{high}}(x)$ denotes high-frequency noise. Another approach is to multiply the Perlin noise texture map with the extinction coefficient ${\beta}_e$ to generate realistic and finely detailed nonuniform smoke patterns.
%

%
%
\begin{table*}[t]
\footnotesize
\renewcommand{\arraystretch}{1.3} 
\setlength{\tabcolsep}{2.8pt} 
\caption{Quantitative comparison of different categories of methods on the SmokeBench dataset.~The best and second-best values are \textbf{blod} and \underline{underlined}.}
\begin{tabularx}{\textwidth}{c|c|ccc|ccc|ccc|ccc|ccc}
\hline
\multirow{2}{*}{Methods} & \multirow{2}{*}{Venue} & \multicolumn{3}{c|}{TestScene -1} & \multicolumn{3}{c|}{TestScene -2} & \multicolumn{3}{c|}{TestScene -3} & 
\multicolumn{3}{c|}{TestScene -4} & 
\multicolumn{3}{c}{Average}\\

& & PSNR$\uparrow$ & SSIM$\uparrow$ & LPIPS$\downarrow$ & PSNR$\uparrow$ & SSIM$\uparrow$ & LPIPS$\downarrow$ & PSNR$\uparrow$ & SSIM$\uparrow$ & LPIPS$\downarrow$ & PSNR$\uparrow$ & SSIM$\uparrow$ & LPIPS$\downarrow$
& PSNR$\uparrow$ & SSIM$\uparrow$ & LPIPS$\downarrow$
\\
 \hline 
FFA-Net~\cite{qin2020ffa} & AAAI'20 & \underline{30.0045}	& 0.9300	& 0.0817 &	\underline{19.1835} &	\underline{0.6615} &	0.4958 &	28.3333 &	0.9272 &	0.1040 &	26.7194 &	\underline{0.8898}	& \underline{0.1193}	& 26.0601	& \underline{0.8521}	& 0.2002
\\ 
MPRNet~\cite{Zamir2021MPRNet} & CVPR'21 & 28.5679	& 0.9154	& 0.0998	& 17.1523	& 0.6223	& 0.5810	& 21.3970	& 0.8958	& 0.1161	& 24.0113	& 0.8709	& 0.1419	& 22.7821	& 0.8261	& 0.2347
\\
Restormer~\cite{Zamir2021Restormer} & CVPR'22 & 29.5115	& 0.9220	& 0.0908	& 18.3266	& 0.6481	& 0.4988	& 26.9610	& 0.9200	& 0.1015	& 25.2410	& 0.8817	& 0.1271	& 25.0100	& 0.8429	& 0.2045
 \\ 
Uformer~\cite{Wang_2022_CVPR} & CVPR'22 & \textbf{30.9107}	& \underline{0.931}	& \underline{0.0791}	& 18.3242	& 0.6554	& 0.4965	& \underline{28.4430}	& \underline{0.9319}	& 0.0907	& \underline{26.8228}	& 0.8885	& \textbf{0.1144}	& \underline{26.1251}	& 0.8517	& 0.1951
\\
DehazeFormer~\cite{song2023vision} & TIP'23 & 29.3786	& 0.9213	& 0.0887	& 18.0116	& 0.6492	& 0.5354	& 22.7881	& 0.9019	& 0.1220	& 24.2829	& 0.8778	& 0.1384	& 23.6153	& 0.8375	& 0.2211
 \\
MB-TaylorFormer~\cite{Yuwei2023MB-TaylorFormer} & ICCV'23 &  29.9594	& \textbf{0.9317}	& \textbf{0.0782}	& \textbf{19.5246}	& \textbf{0.6670}	& \textbf{0.4717}	& \textbf{30.0952}	& \textbf{0.9390}	& \textbf{0.0830}	& \textbf{26.9343}	& \textbf{0.8906}	& 0.1195	& \textbf{26.6283}	& \textbf{0.8570}	& \textbf{0.1881}
  \\ 
X-Restormer~\cite{chen2023comparative} & ECCV'24 & 28.4626	& 0.9153	& 0.0973	& 17.0689	& 0.6331	& 0.5273	& 22.6834	& 0.9006	& 0.1163	& 24.2462	& 0.8750	& 0.1363	& 23.1152	& 0.8310	& 0.2193
\\
MambaIR~\cite{guo2025mambair} & ECCV'24 & 29.5159	& 0.9270	& 0.0827	& 18.6086	& 0.6586	& \underline{0.4898}	& 27.0184	& 0.9290	& \underline{0.0858}	& 25.3036	& 0.8874	& \underline{0.1193}	& 25.1116	& 0.8505	& \underline{0.1944}
\\
 \hline
\end{tabularx}
\label{tab:methods}
\end{table*}

\subsection{Smoke Acquisition System}
To collect a large-scale dataset of real-world smoke image pairs, we develop an early-stage fire scenes smoke acquisition system (SAS), as illustrated in Figure~\ref{fig:data_pipeine}.
The proposed system consists of a surveillance (JOOAN), a smoke machine (Antari Z-1500) and some smoke pellet (Artic Hayes). 
To minimize interference during the data acquisition process, we employ a remote controller for remote operation.
The experimental equipment is shown in Figure~\ref{fig:data_show} (a).
%
{\flushleft\textbf{Scene setup}.}~Our dataset employs a systematic four-dimensional parametric design to ensure complete coverage of real-world fire surveillance scenarios, combining space scale, item density, light condition and shooting angle to construct 43 distinct scenes.  
The data distribution is shown in Figure~\ref{fig:data_show} (b).
The space scale dimension captures variations from large scale rooms exhibiting diluted smoke with ceiling layering to small scale rooms showing rapid saturation, while item density ranges from cluttered density causing strong light reflections and color distortion to sparse single-color backgrounds.
Lighting conditions include natural light with distinct shadows, controlled artificial lights (2700K incandescent and 6000K LED), and IR night vision, with three surveillance shooting angles (high angle, eye level, and low angle) capturing different smoke textures and density distributions. 
This integrated parametric design generates scenarios that represent diverse indoor conditions of the fire scene.
%

To acquire precisely aligned degraded-clean image pairs, we enforces scene controls prior to data collection. 
Specifically, all windows and doors are sealed to create an isolated environment, while lightweight objects are removed to prevent displacement from smoke generator airflow. 
Following scene configuration, absolute stasis of all objects is maintained throughout the recording.

{\flushleft\textbf{Data acquisition}.}~
After finishing the preparatory work, we remotely activate the smoke generator while synchronously recording a complete video of simulated fire smoke conditions through surveillance, from which we extract frames for dataset inclusion. The full acquisition timeline is shown in Figure~\ref{fig:data_pipeine}.

Specifically, before smoke generation, the surveillance system is activated to record the baseline smoke-free phase. 
After starting the smoke generator, the system continuously records the complete dynamic evolution process of smoke diffusion, gradually transitioning from the initial sparse state to the progressively increasing concentration phase, until reaching the saturated state of complete background occlusion when acquisition is terminated.

For dataset construction, the first smoke-free frame serves as the reference ground truth, with subsequent frames uniformly sampled every 2 seconds to systematically capture dynamic smoke state variations at different concentrations, ensuring the acquired data can accurately reflect the time-evolving smoke concentration characteristics in authentic fire scenarios.

{\flushleft\textbf{Data enhancement}.}~
The data enhancement process is structured into three main stages: quality filter, feature augment, and label copy. 
In the quality filter stage, we address the redundancy and lack of information in the initially datasets.  
Since images are captured at uniform intervals from surveillance videos to cover the full smoke generation process, this method often leads to two major issues: at the beginning and end of the smoke event, the smoke concentration changes very slowly or becomes overly dense and uniform.
As a result, many consecutive frames contain little variation or are fully saturated due to excessive smoke accumulation. 
To ensure dataset quality, we manually review and remove such images that exhibit minimal feature differences or are overly saturated, as they are uninformative and prone to causing overfitting during model training.
In the feature augment stage, we address the problem of missing informative frames during the dynamic phase of smoke evolution, where smoke changes rapidly and uniformly timed extraction fails to capture enough diverse samples.  
We manually re-examine the video sequences, identify missed frames that contain unique and informative smoke features, and re-insert them into the dataset to enhance feature diversity.
Finally, we perform label copy to obtain the paired frames, resulting in a final curated dataset comprising 9,975 precisely aligned smoke and smoke-free image pairs, with 9,875 samples allocated for training and 100 reserved for testing and evaluation.
See  Figure~\ref{fig:data_show} (c) for several examples.

\section{Benchmark Evaluation}
\subsection{Evaluation Settings}
{\flushleft\textbf{Comparison baselines}.}~We conduct experiments with eight image desmoking methods on our dataset, including FFA-Net~\cite{qin2020ffa}, MPRNet~\cite{Zamir2021MPRNet} , Restormer~\cite{Zamir2021Restormer}, Uformer~\cite{Wang_2022_CVPR}, DehazeFormer~\cite{song2023vision}, MB-TaylorFormer~\cite{Yuwei2023MB-TaylorFormer}, X-Restormer~\cite{chen2023comparative} and MambaIR~\cite{guo2025mambair}.

{\flushleft\textbf{Implementation details and metrics}.}~To ensure equitable evaluation conditions, all compared methods utilize identical 128×128 patch dimensions, maintaining their respective default configurations for other parameters. The benchmarking is performed using an NVIDIA RTX 4090 GPU, with assessment based on Y-channel PSNR, SSIM, and LPIPS metrics.


\begin{figure*}[!t]
  \centering
   \footnotesize
   \setlength{\tabcolsep}{1.16pt} 
   \renewcommand{\arraystretch}{1} 
  \begin{tabular}{cc}
    \begin{tabular}{c}
      \includegraphics[width=0.353\linewidth]{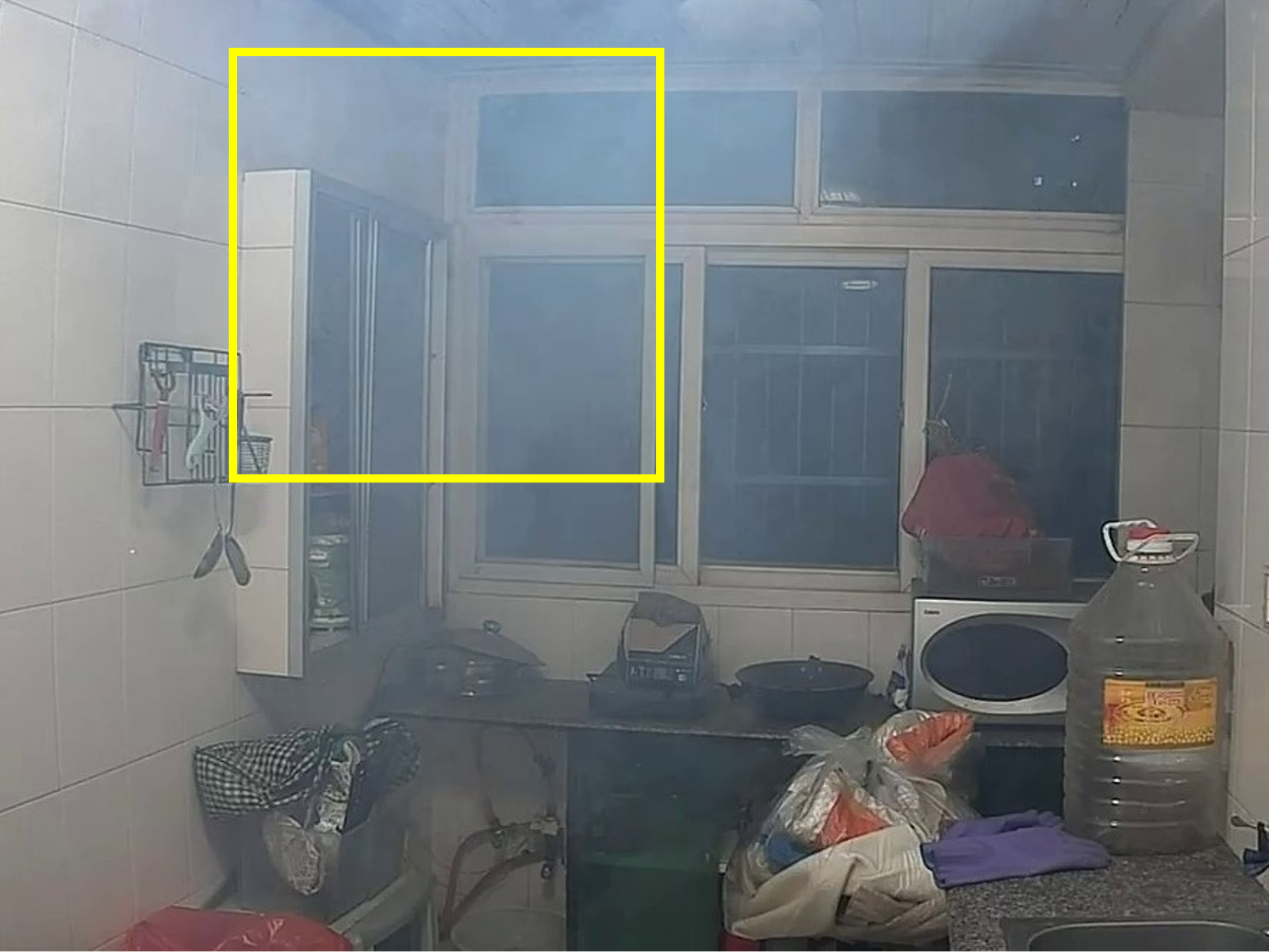} \\
      Thin smoke
    \end{tabular} &
    \begin{tabular}{cccccc}
      \includegraphics[width=0.122\linewidth]{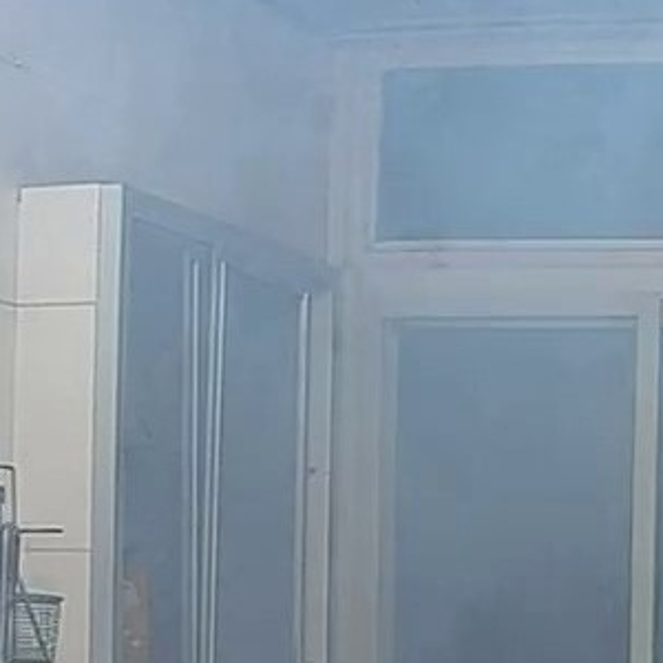} & 
      \includegraphics[width=0.122\linewidth]{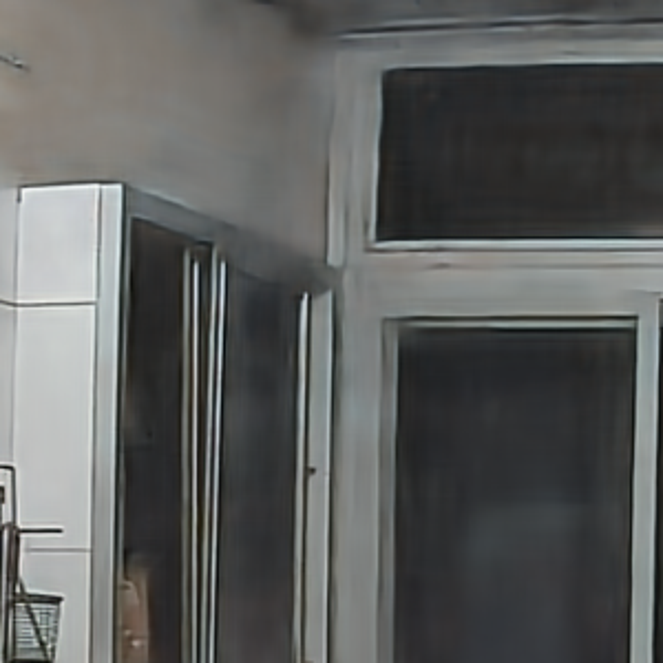} &
      \includegraphics[width=0.122\linewidth]{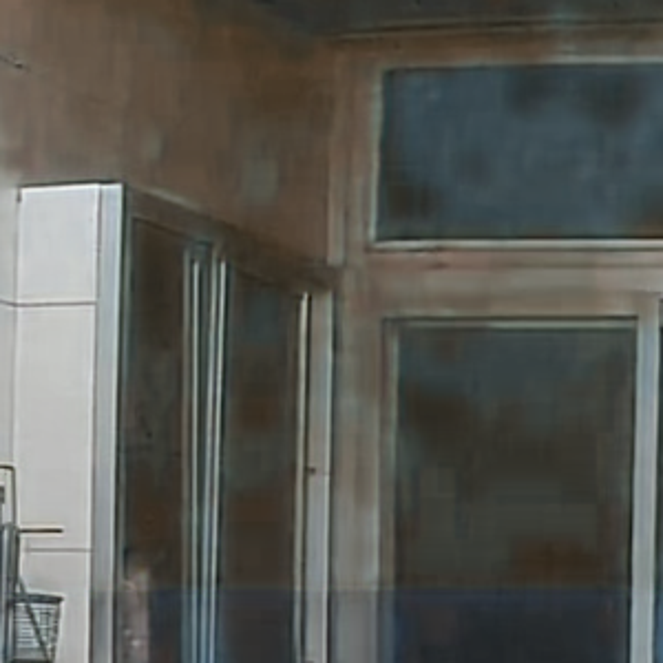} &
      \includegraphics[width=0.122\linewidth]{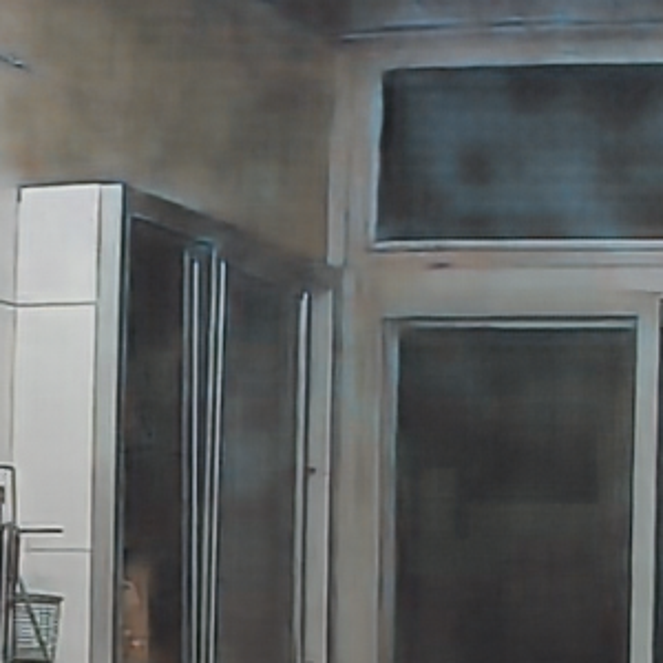} &
      \includegraphics[width=0.122\linewidth]{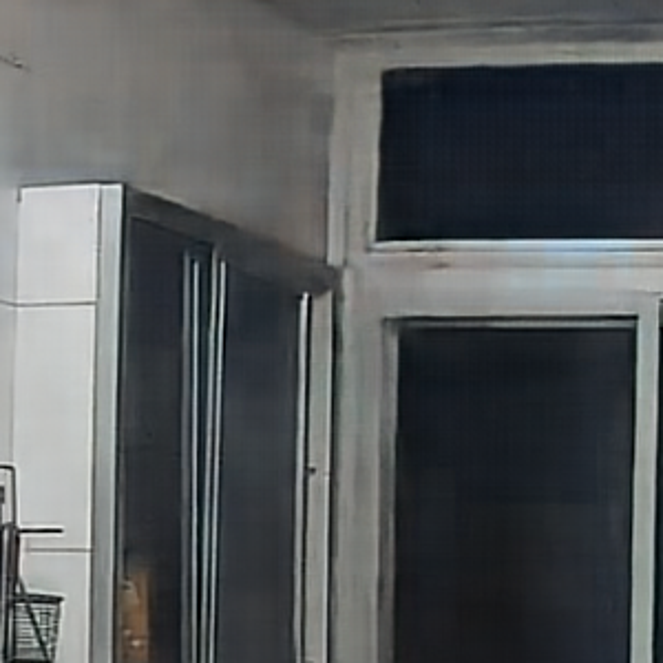} \\
       LQ patch &  FFA-Net~\cite{qin2020ffa} &  MPRNet~\cite{Zamir2021MPRNet} &  Restormer~\cite{Zamir2021Restormer} & Uformer~\cite{Wang_2022_CVPR} \\
      
      \includegraphics[width=0.122\linewidth]{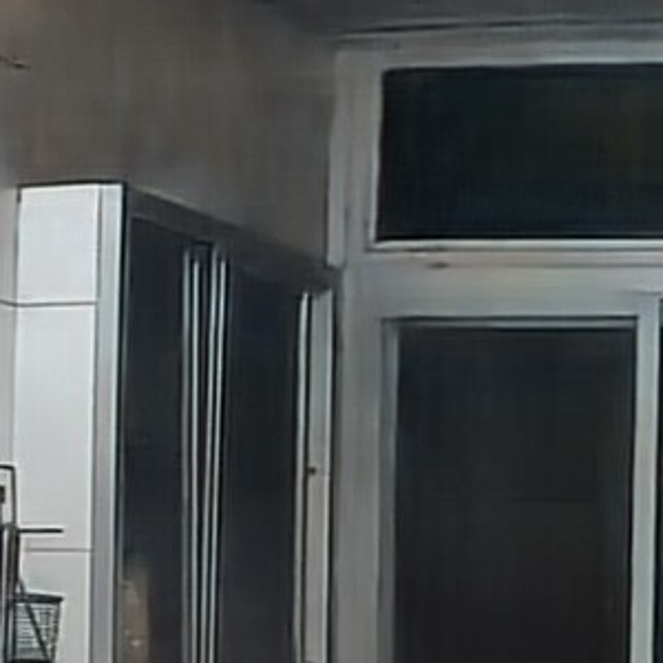} &
      \includegraphics[width=0.122\linewidth]{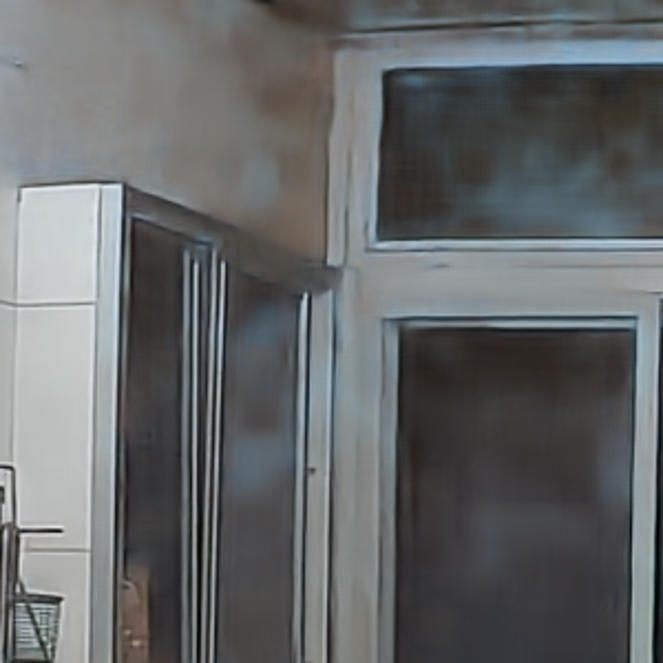} &
      \includegraphics[width=0.122\linewidth]{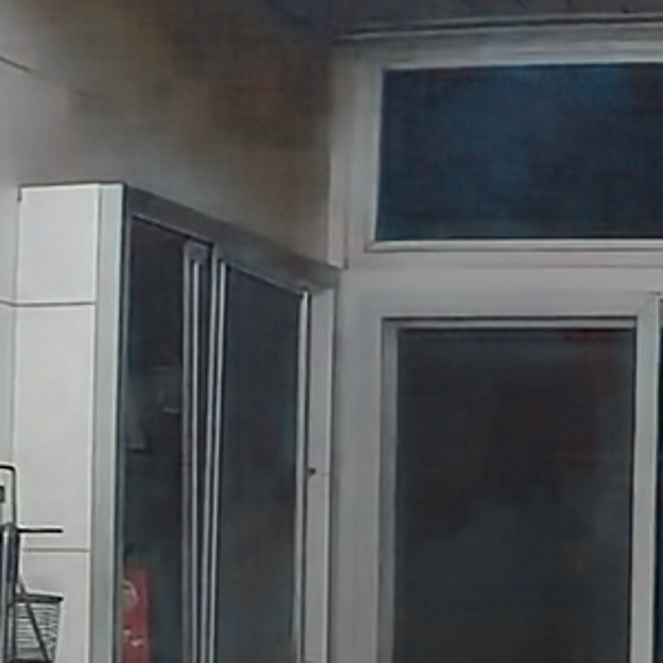} &
      \includegraphics[width=0.122\linewidth]{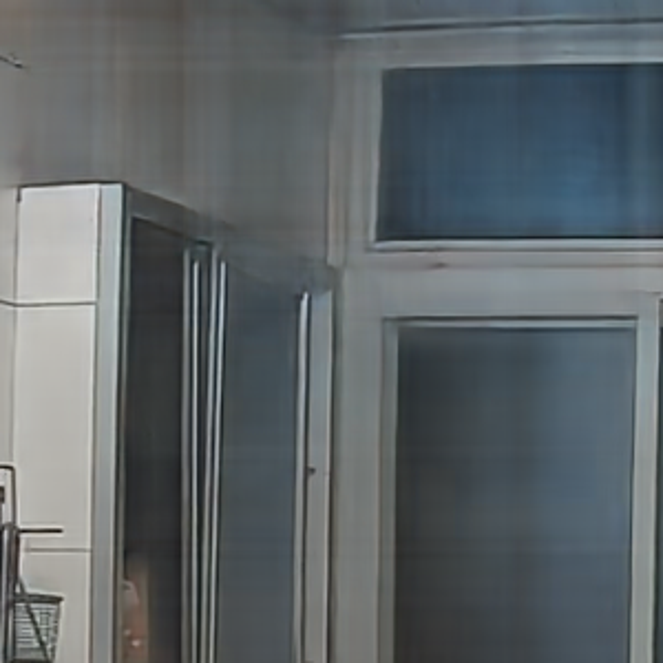} &
      \includegraphics[width=0.122\linewidth]{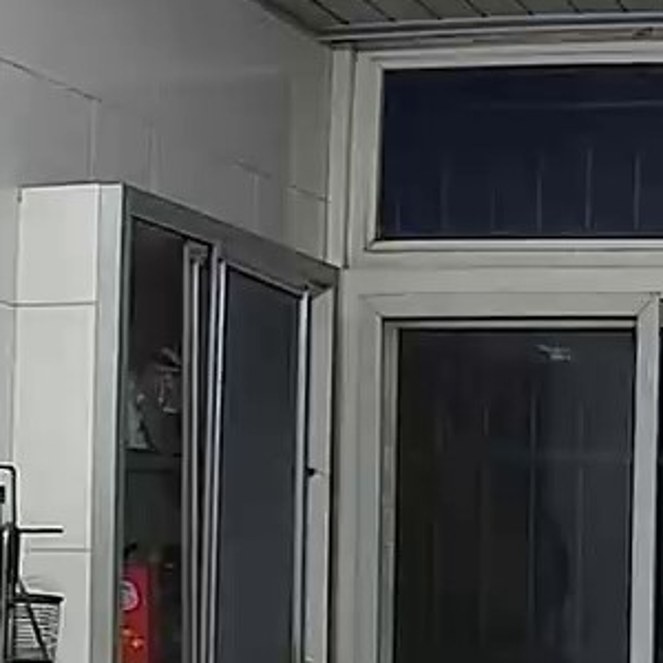} \\
      DehazeFormer~\cite{song2023vision} &TaylorFormer~\cite{Yuwei2023MB-TaylorFormer} & X-Restormer~\cite{chen2023comparative} & MambaIR~\cite{guo2025mambair} &  GT patch
    \end{tabular}
  \end{tabular}


    \begin{tabular}{cc}
    \begin{tabular}{c}
      \includegraphics[width=0.353\linewidth]{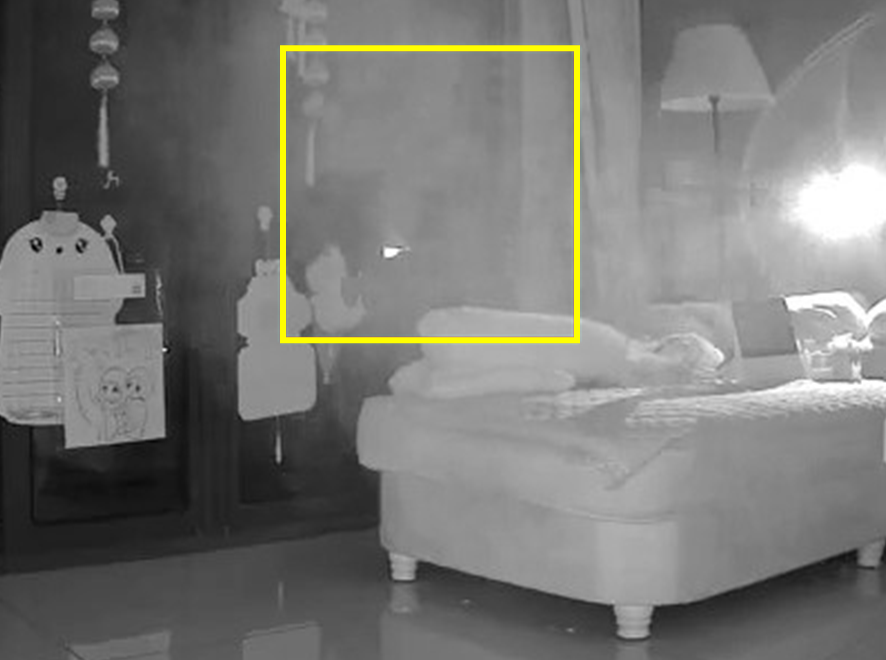} \\
      Medium-density smoke
    \end{tabular} &
    \begin{tabular}{cccccc}
      \includegraphics[width=0.122\linewidth]{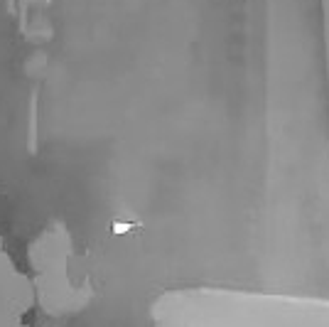} & 
      \includegraphics[width=0.122\linewidth]{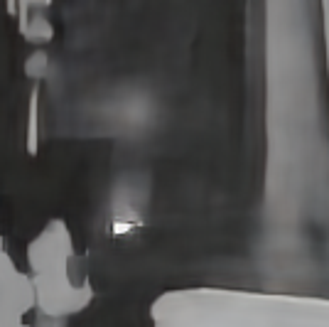} &
      \includegraphics[width=0.122\linewidth]{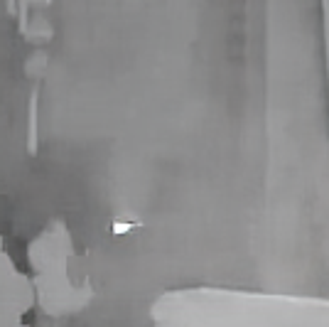} &
      \includegraphics[width=0.122\linewidth]{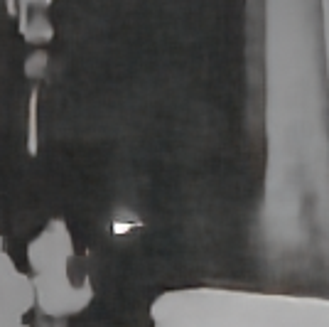} &
      \includegraphics[width=0.122\linewidth]{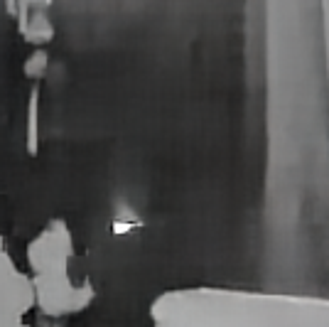} \\
       LQ patch &  FFA-Net~\cite{qin2020ffa} &  MPRNet~\cite{Zamir2021MPRNet} &  Restormer~\cite{Zamir2021Restormer} & Uformer~\cite{Wang_2022_CVPR} \\
      
      \includegraphics[width=0.122\linewidth]{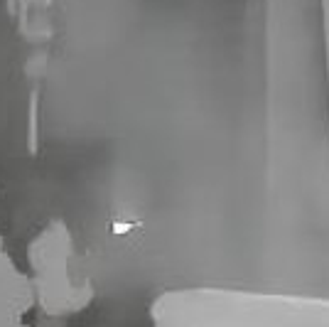} &
      \includegraphics[width=0.122\linewidth]{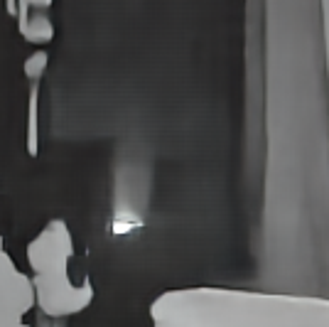} &
      \includegraphics[width=0.122\linewidth]{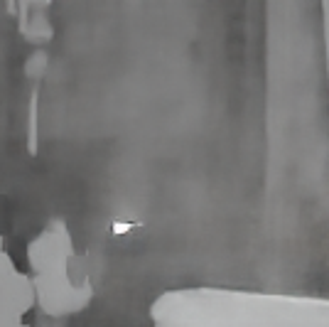} &
      \includegraphics[width=0.122\linewidth]{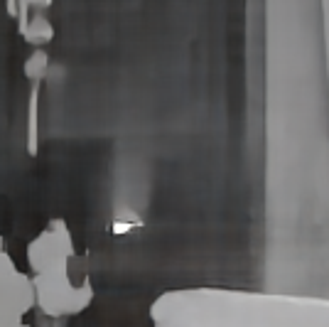} &
      \includegraphics[width=0.122\linewidth]{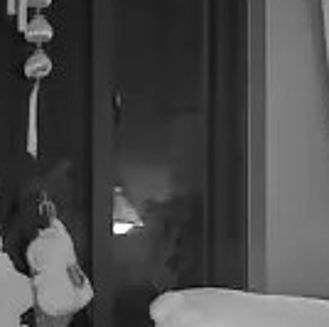} \\
      DehazeFormer~\cite{song2023vision} &TaylorFormer~\cite{Yuwei2023MB-TaylorFormer} & X-Restormer~\cite{chen2023comparative} & MambaIR~\cite{guo2025mambair} &  GT patch
    \end{tabular}
  \end{tabular}


    \begin{tabular}{cc}
    \begin{tabular}{c}
      \includegraphics[width=0.353\linewidth]{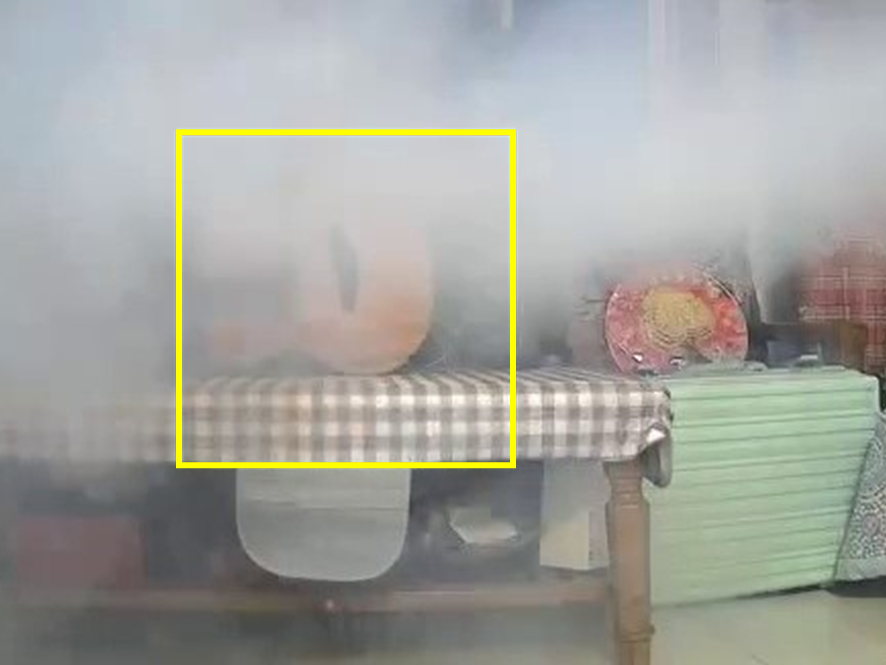} \\
      Thick smoke
    \end{tabular} &
    \begin{tabular}{cccccc}
      \includegraphics[width=0.122\linewidth]{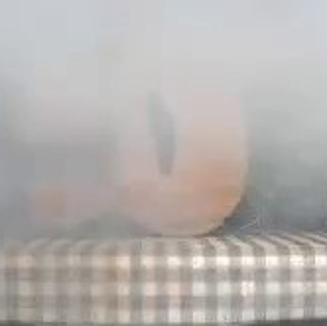} & 
      \includegraphics[width=0.122\linewidth]{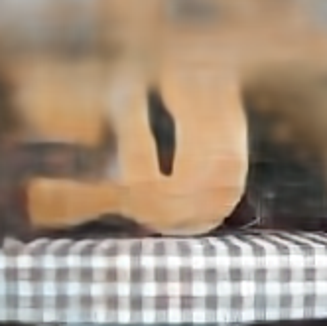} &
      \includegraphics[width=0.122\linewidth]{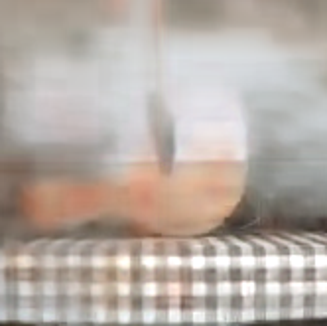} &
      \includegraphics[width=0.122\linewidth]{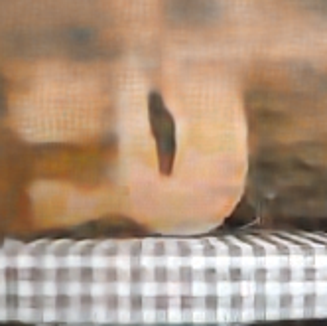} &
      \includegraphics[width=0.122\linewidth]{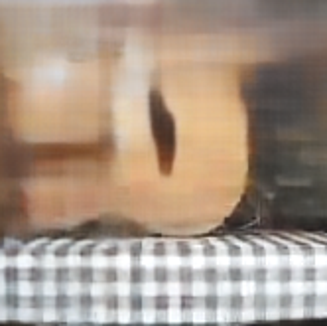} \\
       LQ patch &  FFA-Net~\cite{qin2020ffa} &  MPRNet~\cite{Zamir2021MPRNet} &  Restormer~\cite{Zamir2021Restormer} & Uformer~\cite{Wang_2022_CVPR} \\
      
      \includegraphics[width=0.122\linewidth]{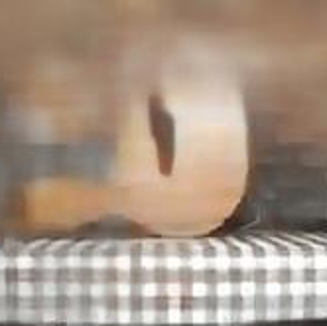} &
      \includegraphics[width=0.122\linewidth]{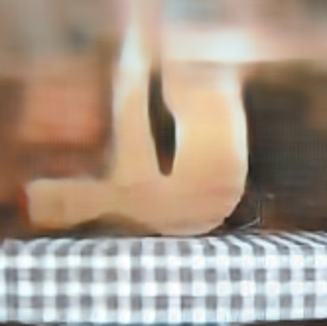} &
      \includegraphics[width=0.122\linewidth]{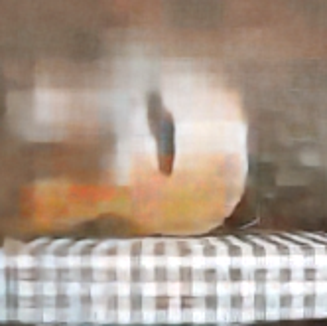} &
      \includegraphics[width=0.122\linewidth]{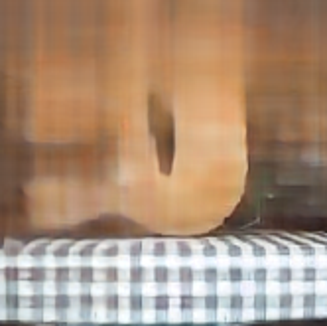} &
      \includegraphics[width=0.122\linewidth]{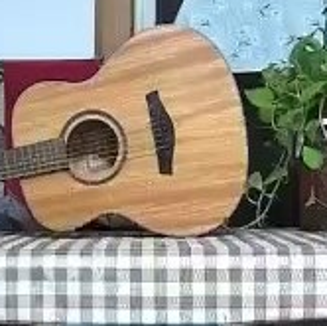} \\
       DehazeFormer~\cite{song2023vision} &TaylorFormer~\cite{Yuwei2023MB-TaylorFormer} & X-Restormer~\cite{chen2023comparative} & MambaIR~\cite{guo2025mambair} &  GT patch
    \end{tabular}
  \end{tabular}

  \caption{Visual comparisons on varying degrees of smoke degradation inputs from the proposed benchmark.
  Left: Overall visualization; Right: Detailed results from different methods. Zoom in for a better view.}
  \label{fig:comparison}
\end{figure*}

\subsection{Quantitative Comparisons}
Table~\ref{tab:methods} presents a quantitative comparison of eight desmoking methods on our dataset, evaluated on four test scenes using PSNR, SSIM, and LPIPS.
MB-TaylorFormer consistently achieves the best overall results, ranking first across all three averaged metrics, highlighting its strong pixel-level and perceptual performance.
Uformer also performs well, achieving the highest SSIM in TestScene-1 and ranking second overall in SSIM, indicating effective structure preservation.
MambaIR shows favorable perceptual quality, ranking second in both SSIM and LPIPS, and excelling in TestScene-3 with the second-best LPIPS.
While methods like Restormer and DehazeFormer yield moderate overall results, they demonstrate strengths in certain scenes, suggesting varied generalization under different smoke densities and scene complexities.
%
\subsection{Qualitative Comparisons}
Figure~\ref{fig:comparison} presents a qualitative comparison of desmoking methods under three levels of smoke density including thin, medium, and thick, providing insight into their visual restoration capabilities beyond numerical metrics.
In the thin smoke scene, most methods succeed in recovering the underlying scene structure. However, differences in color fidelity and texture clarity remain evident. 
For instance, DehazeFormer and Uformer produce relatively sharp and natural-looking results, closely resembling the ground truth, while methods like TaylorFormer and MPRNet tend to leave subtle haze residues or exhibit slight color distortions. 
As the smoke density increases to a medium-density level, disparities between methods become more pronounced. Approaches such as TaylorFormer and MambaIR demonstrate stronger robustness by preserving key semantic features, whereas methods like DehazeFormer and MPRNet struggle with detail recovery, leading to noticeable blurring or incomplete smoke removal.
In the thick smoke scene, none of the evaluated methods can satisfactorily recover plausible structures and textures, with all showing significant artifacts and oversmoothing effects.
%
This fundamental limitation of current approaches strongly motivates our future work to develop more effective desmoking techniques specifically designed for thick smoke conditions.

\section{Conclusion}
%
In this work, we present SmokeBench, the first real-world benchmark dataset specifically designed to address the fundamental challenge of lacking authentic surveillance smoke image data for early-stage fire scene desmoking research.
Unlike methods using synthetic smoke or repurposed dehazing data, our dataset offers authentic image pairs captured under controlled fire conditions, covering diverse smoke densities and scenes.
Publicly released to support community research efforts, SmokeBench's dataset and acquisition framework enable groundbreaking future advances in both multi-modal fire imaging and real-time edge deployment for intelligent emergency response systems.



\bibliographystyle{ACM-Reference-Format}
\balance
\bibliography{sample-base}










\end{document}